%% file: main.tex
\documentclass[a4,twocolumn]{article}

\usepackage[utf8]{inputenc}
\usepackage[usenames,dvipsnames]{xcolor}
\usepackage{graphicx} 
\usepackage{subfigure} 

\usepackage{natbib}

\usepackage[ruled, linesnumbered]{algorithm2e}

\usepackage{amssymb,amsmath}

\usepackage{hyperref}
\definecolor{mydarkblue}{rgb}{0,0.08,0.45}
\hypersetup{ %
pdftitle={},
pdfauthor={},
pdfsubject={Proceedings of the International Conference on Machine Learning 2010},
pdfkeywords={},
pdfborder=0 0 0,
pdfpagemode=UseNone,
colorlinks=true,
linkcolor=mydarkblue,
citecolor=mydarkblue,
filecolor=mydarkblue,
urlcolor=mydarkblue,
pdfview=FitH}

\DeclareMathOperator*{\argmax}{argmax}
\DeclareMathOperator*{\argmin}{argmin}
\newcommand{\numberofactions}{A}
\newcommand{\numstates}{S}
\newcommand{\numactions}{A}
\newcommand{\numrollouts}{K}
\newcommand{\numclassifiers}{C}
\newcommand{\numclassifierslog}{\gamma \log(A)}
\newcommand{\trajlength}{T}
\newcommand{\state}{s}
\newcommand{\action}{a}
\newcommand{\stateset}{\mathcal{S}}
\newcommand{\actionset}{\mathcal{A}}
\newcommand{\trainingset}{\mathcal{S}_\text{T}}
\newcommand{\transitionfunc}{T}
\newcommand{\rewardfunc}{R}
\newcommand{\codingmatrix}{\mathbf{M}^{c}}
\newcommand{\mdp}{\mathcal{M}}
\newcommand{\bigO}{\mathcal{O}}
\usepackage{authblk}

\begin{document} 

\title{Fast Reinforcement Learning with Large Action Sets using Error-Correcting Output Codes for MDP Factorization}

\author[1]{Gabriel Dulac-Arnold}
\author[1]{Ludovic Denoyer}
\author[2]{Philippe Preux}
\author[1]{Patrick Gallinari}
\affil[1]{UPMC--LIP6, Case 169 -- 4 Place Jussieu, Paris 75005, France\footnote{firstname.lastname@lip6.fr}}
\affil[2]{LIFL (UMR CNRS) \& INRIA Lille Nord-Europe -- Université de Lille, Villeneuve d'Ascq, France\footnote{firstname.lastname@inria.fr}}
\date{\today}

\maketitle
\begin{abstract} 
The use of Reinforcement Learning in real-world scenarios is strongly limited by issues of scale.  
Most RL learning algorithms are unable to deal with problems composed of hundreds or sometimes even dozens of possible actions, and therefore cannot be applied to many real-world problems. We consider the RL problem in the supervised classification framework where the optimal policy is obtained through a multiclass classifier, the set of classes being the set of actions of the problem.
We introduce error-correcting output codes (ECOCs) in this setting and propose two new methods for reducing complexity when using rollouts-based approaches. The first method consists in using an ECOC-based classifier as the multiclass classifier, reducing the learning complexity from $\mathcal{O}(\numberofactions^2)$ to $\mathcal{O}(\numberofactions \log(\numberofactions))$. 
We then propose a novel method that profits from the ECOC's coding dictionary to split the initial MDP into $\mathcal{O}(\log(\numberofactions))$ seperate two-action MDPs. This second method reduces learning complexity even further, from $\mathcal{O}(\numberofactions^2)$ to $\mathcal{O}(\log(\numberofactions))$, thus rendering problems with large action sets tractable.
We finish by experimentally demonstrating the advantages of our approach on a set of benchmark problems, both in speed and performance. 
\end{abstract} 

\section{Introduction}
\label{sec:intro}
\input{sections/intro}

\section{Background}
\label{sec:background}
\input{sections/background}

\section{Extended \& Binary RCPI}
\label{sec:contributions}
\input{sections/contributions}

\input{sections/complexity}

\section{Experiments}
\label{sec:experiments}
\input{sections/experiments}

\section{Related Work}
\label{sec:related}
\input{sections/related}

\section{Conclusion}
\label{sec:conclusion}
\input{sections/conclusion}

\bibliography{./ICML2012}
\bibliographystyle{icml2012}
\end{document}

%% file: sections/intro.tex
The goal of Reinforcement Learning (RL) and more generally sequential decision making is to learn an optimal policy for performing a certain task within an environment, modeled by a Markov Decision Process (MDP). In RL, the dynamics of the environment are considered as unknowns. This means that to obtain an optimal policy, the learner interacts with its environment, observing the outcomes of the actions it performs.
Though well understood from a theoretical point of view, RL
still faces many practical issues related to the complexity of the environment, in particular when dealing with large state or action sets.
Currently, using function approximation to better represent and generalize over the environment is a common approach for dealing with large \textit{state} sets. 
However, learning with large \textit{action} sets has been less explored and remains a key challenge.
 
When the number of possible actions $\numactions$ is neither on the scale of `a few' nor 
outright continuous, the situation becomes difficult. 
In particular cases where the action space is continuous (or nearly so),
a regularity assumption can be made on the consequences of the actions concerning either a certain smoothness  or Lipschitz property over the action space \cite{Lazaric2007, bubeck2011x, negoescu2010knowledge}. 
However, situations abound in which the set of actions is discrete, but the number of actions lies somewhere between 10 and $10^4$ (or greater) --- Go, Chess, and planning problems are among these.  
In the common case where the action space shows no regularity, it is not possible to gain knowledge regarding the consequence of an action that has never been applied --- sub-sampling is therefore not an option.

In this article, we present an algorithm which can intelligently sub-sample even completely irregular action spaces.
Drawing from ideas used in multiclass supervised learning, \textbf{we introduce a novel way to significantly reduce the complexity of learning (and acting) with large action sets}.  By assigning a multi-bit code to each action, we create binary clusters of actions through the use of \textit{Error Correcting Output Codes} (ECOCs) \cite{Dietterich1995}.  
Our approach is anchored in Rollout Classification Policy Iteration (RCPI) \cite{Lagoudakis2003}, an algorithm well know for its efficiency on real-world problems.  
We begin by proposing a simple way to reduce the computational cost of any policy by leveraging the clusters of actions defined by the ECOCs.
We then extend this idea to the problem of learning, and propose a new RL method that allows one to find an approximated optimal policy by solving a set of 2-action MDPs.
While our first model --- ECOC-extended RCPI (ERCPI) --- reduces the overall learning complexity from $\mathcal{O}(\numactions^2)$ to $\mathcal{O}(\numactions \log(\numactions))$, our second method --- Binary-RCPI (BRCPI) ---  reduces this complexity even further, to $\mathcal{O}(\log(\numactions))$.  

The paper is organized as follow: We give a brief overview of notation and RL in Section \ref{sec:background_rl}, then introduce RCPI and ECOCs in Sections \ref{sec:background_rcpi} and \ref{sec:background_ecoc} respectively.  We present the general idea of our work in Section \ref{sec:contributions}. We show how RCPI can be extended using ECOCs in \ref{sec:contrib_ecoc}, and then explain in detail how an MDP can be factorized to accelerate RCPI during the learning phase in \ref{sec:BRCPI}. An in-depth complexity analysis of the different algorithms is given in Section \ref{sec:complexity}. Experimental results are provided on two problems in Section \ref{sec:experiments}. Related work is presented in Section \ref{sec:related}.

%% file: sections/background.tex
In this section, we cover the three key elements to understanding our work: Markov Decision Problems, Rollout Classification Policy Iteration, and Error-Correcting Output Codes.

\subsection{Markov Decision Process}
\label{sec:background_rl}
Let a Markov Decision Process $\mdp$ be defined by a 4-tuple $\mdp = (\stateset, \actionset, \transitionfunc, \rewardfunc)$.
\begin{itemize}
\item  $\stateset$ is the set of possible states of the MDP, where $\state \in \stateset$ denotes one state of the MDP.
 \item $\actionset$ is the set of possible actions, where $\action \in \actionset$ denotes one action of the MDP.
 \item  $\transitionfunc : \stateset \times \stateset \times \actionset \rightarrow \mathbb{R}$ is the MDP's transition function, and defines the probability of going from state $\state$ to state $\state'$ having chosen action $\action$: $T(\state',\state,\action) = P(\state' | \state, \action)$.
 \item $\rewardfunc : \stateset \times \actionset \rightarrow \mathbb{R}$ is a reward function defining the expected immediate reward of taking action $\action$ in state $\state$.  
 The actual immediate reward for a particular transition is denoted by $r$.
\end{itemize}

In this article, we assume that the set of possible actions is the same for all states, but our work is not restricted to this situation; the set of actions can vary with the state without any drawbacks.

Let us define a policy, $\pi: \stateset \rightarrow \actionset$, providing a mapping from states to actions in the MDP. 
In this paper, without loss of generality, we consider that the objective to fulfill is the optimization of the expected sum of $\gamma$-discounted rewards from a given set of states $D$: $J_\pi(s) = \mathbb{E} [\sum_{k\ge{}0} \gamma^k r_{t+k} | s_t = s \in D, \pi]$.

A policy's performance is measured w.r.t. the objective function $J_\pi$.
The goal of RL is to find an optimal policy $\pi^*$ that maximizes the objective function: $\pi^* = arg max_\pi J_\pi$.

In an RL problem, the agent knows both $\stateset$ and $\actionset$, but is not given the environment's dynamics defined by $\transitionfunc$ and $\rewardfunc$.
In the case of our problems, we assume that the agent may start from any state in the MDP, and can run as many simulations as necessary until it has learned a good policy.

\subsection{Rollout Classification Policy Iteration}
\label{sec:background_rcpi}
We anchor our contribution in the framework provided by RCPI \cite{Lagoudakis2003}.
RCPI belongs to the family of Approximate Policy Iteration (API) algorithms, iteratively improving estimates of the $Q$-function --- $Q_\pi (s, a) = \mathbb{E} [J_\pi (s) | \pi]$.
In general, API uses a policy $\pi$ to estimate $Q$ through simulation, and then approximates it by some form of regression on the estimated values, providing $\tilde{Q}_\pi$.
This is done first with an initial (and often random) policy $\pi_0$, and is iteratively repeated until $\tilde{Q}_\pi$ is properly estimated.
$\tilde{Q}_\pi(\state, \action)$ is estimated by running $\numrollouts$ \textit{rollouts} i.e. Monte-Carlo simulations using $\pi$ to estimate the expected reward. The new policy $\pi'$ is thus the policy that chooses the action with the highest $\tilde{Q}_\pi$-value for each state.

In the case of RCPI, instead of using a function approximator to estimate $\tilde{Q}_\pi$, the best action for a given $s$ is selected using a classifier, without explicitly approximating the Q-value. This estimation is usually done using a binary classifier $f_\theta$ over the state-action space such that the new policy can be written as:
\begin{equation}
\label{eq:rcpi_argmax}
\pi'(s) = \argmax_{\action \in \actionset_s} f_\theta(\state, \action).
\end{equation}
The classifier's training set $\trainingset$ is generated through Monte-Carlo sampling of the MDP, estimating the optimal action for each state sampled.  Once generated, these optimal state-action pairs $(\state, \action)$ are used to train a supervised classifier; the state is interpreted as the feature vector, and the action $\action$ as the state's label.
In other words, RCPI is an API algorithm that uses Monte Carlo simulations to transform the RL problem into a multiclass classification problem.


\subsection{Error-Correcting Output Codes}
\label{sec:background_ecoc}

In the domain of multiclass supervised classification in large label spaces, ECOCs have been in use for a while \cite{Dietterich1995}. We will cover ECOCs very briefly here, as their adaptation to an MDP formalism is well detailed in Section \ref{sec:contrib_ecoc}.

Given a multiclass classification task with a label set $\mathcal{Y}$, the $|\mathcal{Y}|$ class labels can be encoded as binary integers using as few as $\numclassifiers =  \log_2(|\mathcal{Y}|)$ bits. ECOCs for classification assume that each label is associated to a binary code of length\footnote{Different methods exist for generating such codes. In practice, it is customary to use redundant codes where $\gamma \approx 10$.} $C=\gamma \log(|\mathcal{Y}|)$ with $\gamma \geq 1$. 

The main principle of multiclass classifiers with ECOCs is to learn to predict the output code instead of directly predicting the label, transforming a supervised learning problem with $|\mathcal{Y}|$ classes into a set of $C=\gamma \log(|\mathcal{Y}|)$ binary supervised learning problems. 
Once trained, the class of a datum $x$ can be inferred by passing the datum to all the classifiers and concatenating their output into a predicted label code: $code(x) = (f_{\theta_0}(x), \cdots, f_{\theta_C}(x))$. The predicted label is thus the label with the closest code in terms of Hamming distance. As a side note, Hamming distance look-ups can be done in logarithmic time by using tree-based approaches such as \textit{k}-d trees \cite{Bentley1975}.  ECOCs for classification can thus infer with a complexity of $\bigO(log(|\mathcal{Y}|)$.

%% file: sections/contributions.tex
We separate this paper's contributions into two parts, the second part building on the first one.  
We begin by showing how ECOCs can be easily integrated into a classifier-based policy, and proceed to show how the ECOC's coding matrix can be used to factorize RCPI into a much less complex learning algorithm.

\subsection{General Idea}

\begin{figure}	
\small{
  \begin{center}
    \begin{tabular}{| c || c | c | c | } \hline
       & $b_1$ & $b_2$ & $b_3$ \\ 
      \hline \hline
      $\action_1$ & $+$ & $+$ & $-$ \\
      $\action_2$ & $-$ & $+$ & $-$ \\
      $\action_3$ & $+$ & $-$ & $+$ \\
      $\action_4$ & $-$ & $+$ & $+$ \\
      $\action_5$ & $+$ & $-$ & $-$ \\
      \hline  
    \end{tabular}
  \end{center}
 }
  \caption{An example of a 5-actions, $\numclassifiers=3$-bits coding matrix. The code of action $1$ is $(+,+,-)$.} 
  \label{fig:codingmatrix}
\end{figure}

The general idea of our two algorithms revolves around the use of ECOCs for representing the set of possible actions, $\actionset$. This approach assigns a multi-bit code of length $\numclassifiers = \gamma\log(\numactions)$ to each of the $\numactions$ actions. The codes are organized in a \textit{coding matrix}, illustrated in Figure~\ref{fig:codingmatrix} and denoted $\codingmatrix$.  Each row corresponds to one action's binary code, while each column is a particular dichotomy of the action space corresponding to that column's associated bit $b_i$. In effect, each column is a projection of the $\numactions$-dimensional action space into a 2-dimensional binary space. We denote $\codingmatrix_{[\action,*]}$ as the $\action^{th}$ row of $\codingmatrix$, which corresponds to $\action$'s binary code.  $\codingmatrix_{[\action,i]}$ corresponds to bit $b_i$ of action $\action$'s binary code.

\textbf{Our main idea is to consider that each bit corresponds to a binary sub-policy denoted} $\pi_i$.  
By combining these sub-policies, we can derive the original policy $\pi$ one wants to learn as such:
 \begin{equation}
   \label{eq:ecoc_policy}
   \pi(s) = \argmin_{\action \in \actionset} d_{\text{H}}(\mathbf{M}^{c}_{[\action, *]}, (\pi_1(s), \cdots, \pi_\numclassifiers(s)),
 \end{equation}
where $\mathbf{M}^{c}_{[\action, *]}$ is the binary code for action $\action$, and $d_{\text{H}}$ is the Hamming distance. 
For a given a state $\state$, each sub-policy provides a binary action $\pi_i (\state) \in \{ -, + \}$, thus producing a binary vector of length $\numclassifiers$.   $\pi(s)$ chooses the action $\action$ with the binary code that has the smallest Hamming distance to the concatenated output of the $\numclassifiers$ binary policies.

We propose two variants of RCPI that differ by the way they learn these sub-policies. ECOC-extended RCPI (ERCPI) replaces the standard definition of $\pi$ by the definition in Eq. \eqref{eq:ecoc_policy}, both for learning and action selection.
The Binary-RCPI method (BRCPI) relaxes the learning problem and considers that all the sub-policies can be learned independently on separate binary-actioned MDPs, resulting in a very rapid learning algorithm.

\subsection{ECOC-Extended RCPI}
\label{sec:contrib_ecoc}

ERCPI takes advantage of the policy definition in Equation \eqref{eq:ecoc_policy} to decrease RCPI's complexity. 
The $\numclassifiers$ sub-policies --- $\pi_{i \in [1,\numclassifiers]}$ --- are learned simultaneously on the original MDP, by extending the RCPI algorithm with an ECOC-encoding step, as described in Algorithm \ref{alg:rcpi_ecoc}.
As any policy improvement algorithm, ERCPI iteratively performs the following two steps:
 
  \textbf{Simulation Step}: This consists in performing Monte Carlo simulations to estimate the quality of a set of state-action pairs. From these simulations, a set of training examples $\trainingset$ is generated, in which data are states, and labels are the estimated best action for each state. 
  
  \textbf{Learning Step}:  For each bit $b_i$, $\trainingset$ is used to create a binary label training set $\trainingset^i$. Each $\trainingset^i$ is then used to train a classifier $f_{\theta_i}$, providing sub-policy $\pi_i'$ as in Eq. \eqref{eq:rcpi_argmax}. Finally, the set of $\numclassifiers$ sub-policies are combined to provide the final improved policy as in Eq. \eqref{eq:ecoc_policy}.

ERCPI's training algorithm is presented in Alg. \ref{alg:rcpi_ecoc}.

The \textbf{Rollout} function used by ERCPI is identical to the one used by RCPI --- $\pi$ is used to estimate a certain state-action tuple's expected reward, $\tilde{Q}_\pi(\state,\action)$. 

\begin{algorithm}[ht]
\footnotesize{
\label{alg:rcpi_ecoc}
\KwData{\\
$\stateset_R$: uniformly sampled state set; 
$\mathcal{M}$: MDP; 
$\pi_0$: initial policy; 
$\numrollouts$: number of trajectories; 
$\trajlength$: maximum trajectory length\\
}
$\pi = \pi_0$\\
\Repeat{$\pi\sim\pi'$}{
$\trainingset=\emptyset$\\
 \ForEach{$s \in \stateset_R$}{
    \ForEach{$a \in A$}{

       $\tilde{Q}_\pi(\state,\action) \leftarrow \mathbf{Rollout}(\mathcal{M},\state,\action,\numrollouts,\pi)$\\ 
  }
  $\action^* = \argmax_{\action \in \actionset} \tilde{Q}_\pi(\state,\action) $\\
  \If{$\forall \action \neq \action^*, \tilde{Q}_\pi(\state,\action) \ll \tilde{Q}_\pi(\state,\action^*)$}
  {
  $\trainingset = \trainingset \cup \{(\state,\action^*)\}$\\
  }
  
  } 
  \ForEach{$i \in [1,\numclassifiers]$}{
    $\trainingset^i = \emptyset$ \\
    \ForEach{$(\state, \action) \in \trainingset$}{
    $\action_i = \codingmatrix_{[\action, i]}$\\
    $\trainingset^i = \trainingset^i \cup (\state, \action_i)$
    }
    $f_{\theta_i} = \mathbf{Train}(\trainingset^i)$\\
    $\pi'_i$ from $f_{\theta_i}$ as defined in Eq. \eqref{eq:rcpi_argmax}\\
  }
 $\pi'$ as defined in Eq. \eqref{eq:ecoc_policy}\\
 $\pi = \alpha(\pi,\pi')$
}
\Return{ $\pi$}
}
\caption{ERCPI}
\end{algorithm}
Up to line 12 of Algorithm \ref{alg:rcpi_ecoc}, ERCPI is in fact algorithmically identical to RCPI, with the slight distinction that only the best $(\state,\action^*)$ tuples are kept, as is usual when using RCPI with a multiclass classifier.

ERCPI's main difference appears starting line 13; it is here that the original training set $\trainingset$ is mapped onto the $\numclassifiers$ binary action spaces, and that each individual sub-policy $\pi_i$ is learned. Line 16 replaces the original label of state $\state$ by its binary label in $\pi_i$'s action space --- this corresponds to bit $b_i$ of action $\action$'s code.

The \textbf{Train} function on line 19 corresponds to the training of $\pi_i$'s corresponding binary classifier on $\trainingset^i$.  After this step, the global policy $\pi'$ is defined according to Eq.\eqref{eq:ecoc_policy}. Note that, to ensure the stability of the algorithm, the new policy $\pi$ obtained after one iteration of the algorithm is an alpha-mixture policy between the old $\pi$ and the new $\pi'$ obtained by the classifier (cf. line 23).

\subsection{Binarized RCPI}
\label{sec:BRCPI}
ERCPI splits the policy improvement problem into $\numclassifiers$ individual problems, but training still needs $\pi$, thus requiring the full set of binary policies.  Additonnally, for each state, all $\numactions$ actions have to be evaluated by Monte Carlo simulation (Alg. \ref{alg:rcpi_ecoc}, line 5). To reduce the complexity of this algorithm, we propose learning the $\numclassifiers$ binary sub-policies --- $\pi_{i \in [1,\numclassifiers]}$ --- \textbf{independently}, transforming our initial MDP into $\numclassifiers$ sub-MDPs, each one corresponding to the environment in which a particular $\pi_i$ is acting.

Each of the $\pi_i$ binary policy is dealing with its own particular representation of the action space, defined by its corresponding column in $\codingmatrix$.  For training, best-action selections must be mapped into this binary space, and each of the $\pi_i$'s choices must be combined to be applied back in the original state space.  

Let $\actionset^+_i,\actionset^-_i \subset \actionset$ be the action sets associated to $\pi_i$ such that:
\begin{equation}
\begin{aligned}
\actionset^+_i & = \{\action \in \actionset ~|~ \codingmatrix_{[\action, i]} = \text{`}+\text{'}\} \\
\actionset^-_i &= \actionset \setminus \actionset^+_i = \{\action \in \actionset ~|~ \codingmatrix_{[\action, i]} = \text{`}-\text{'}\}.
\end{aligned}
\end{equation}
For a particular $i$, $\actionset^+_i$ is the set of original actions corresponding to sub-action $+$, and $\actionset^-_i$ is the set of original actions corresponding to sub-action $-$. 

We can now define $\numclassifiers$ new binary MDPs that we name sub-MDPs, and denote $\mathcal{M}_{i \in [1,\numclassifiers]}$. They are defined from the original MDP as follows:
\begin{itemize}
 \item $\stateset_i = \stateset$, the same state-set as the original MDP.
 \item $\actionset_i = \{+,-\}$.
 \item $\transitionfunc_i = \transitionfunc(s',s,a)P(a|a_i) = P(s'|s,a)P(a|a_i)$, where $P(a|a_i)$ is the probability of choosing action $a\in \actionset^{a_i}$, knowing that the sub-action applied on the sub-MDP $\mathcal{M}_i$ is $a_i \in \{+,-\}$. We consider $P(a|+)$ to be uniform for $\action \in \actionset^+$ and null for $\action \in \actionset^-$, and vice versa. $P(s'|s,a)$ is the original MDP's transition probability. 
 \item $\rewardfunc_i(s,a_i) = \sum\limits_{a \in \actionset^{a_i}_i}P(a|a_i)\rewardfunc(s,a)$. 
\end{itemize}

Each of these new MDPs represents the environment in which a particular binary policy $\pi_i$ operates.  Each of these MDPs is defined independently from one another, and therefore we can consider each of these MDPs to be a separate RL problem for its corresponding binary policy.

In light of this, we propose to transform RCPI's training process for the base MDP into $\numclassifiers$ new training processes, each one trying to find an optimal $\pi_i$ for its corresponding $\mathcal{M}_i$.  Once all of these binary policies have been trained, they can be used during inference in the manner described in Section \ref{sec:contrib_ecoc}.

The main advantage of this approach is that, since each of the $\numclassifierslog$ sub-problems in Algorithm \ref{alg:rcpi_binary} is modeled as a binary-actioned MDP, increasing the number of actions in the original problem simply increases the number of sub-problems logarithmically, without increasing the complexity of these sub-problems -- see Section \ref{sec:complexity}.

\begin{algorithm}[h]
\begin{footnotesize}
\label{alg:rcpi_binary}
\KwData{\\
$\stateset_R$: uniformly sampled state set; 
$\mathcal{M}$: MDP; 
$\pi_0$: random policy; 
$\numrollouts$: number of trajectories; 
$\trajlength$: maximum trajectory length;
$\numclassifiers$: number of binary MDPs\\
}
\ForEach{$i \in \numclassifiers$}{
$\pi_i = \pi_0$\\
\Repeat{$\pi_i \sim \pi_i'$}{
$\trainingset=\emptyset$\\
\ForEach{$s \in \stateset_R$}{
  \ForEach{$a \in \{+,-\}$}{
       $\tilde{Q}_\pi(\state,\action) \leftarrow \mathbf{Rollout}(\mathcal{M}_i,\state,\action,\numrollouts,\pi_i)$\\ 
  }
  $\action^* = \argmax_{\action \in \actionset} \tilde{Q}_\pi(\state,\action) $\\
  \If{$\forall \action \neq \action^*, \tilde{Q}_\pi(\state,\action) \ll \tilde{Q}_\pi(\state,\action^*)$}
  {
  $\trainingset = \trainingset \cup \{(\state,\action^*)\}$\\
  }
  } 
  $f_{\theta_i} = \mathbf{Train}(\trainingset)$\\

 $\pi_i'$ from $f_{\theta_i}$ as defined in Eq. \eqref{eq:rcpi_argmax}\\
 $\pi_i = \alpha(\pi_i,\pi_i')$
}
\Return{$\pi$ as defined in Eq. \eqref{eq:ecoc_policy}}
}
\caption{BRCPI}
\end{footnotesize}
\end{algorithm}

Let us now discuss some details of BRCPI, as described in Algorithm \ref{alg:rcpi_binary}.  BRCPI resembles RCPI very strongly, except that instead of looping over the $\numactions$ actions on line 6, BRCPI is only sampling $\tilde{Q}$ for $+$ or $-$ actions.  However, the inner loop is run $\numclassifiers = \numclassifierslog$ times, as can be seen on line 1 of  Algorithm \ref{alg:rcpi_binary}.

Within the \textbf{Rollout} function (line 7), if $\pi_i$ chooses sub-action `$+$', an action $\action_i$ from the original MDP is sampled from $\actionset^+_i$ following $P(a|a_i)$, and the MDP's transition function is called using this action.  This effectively estimates the expected reward of choosing action $+$ in state $\state$.

As we saw in Section \ref{sec:contrib_ecoc}, each $\actionset_i$ is a different binary projection of the original action set.  Each of the $\pi_i$ classifiers is thus making a decision considering a different split of the action space.  Some splits may make no particular sense w.r.t. to the MDP at hand, and therefore the expected return of that particular $\pi_i$'s $\actionset^+_i$ and $\actionset^-_i$ may be equal.  This does not pose a problem, as that particular sub-policy will simply output noise, which will be corrected for by more pertinent splits given to the other sub-policies.

%% file: sections/complexity.tex
\subsection{Computational Cost and Complexity}
\label{sec:complexity}

We study the computational cost of the proposed algorithms in comparison with the RCPI approach and present their respective complexities.

In order to define this cost, let us consider that  $\mathcal{C}(\numstates,\numactions)$ is the time spent learning a multiclass classifier on $\numstates$ examples with $\numactions$ possible outputs, and $\mathcal{I}(\numactions)$ is the cost of classifying one input. 

The computational cost of one iteration of RCPI or ERCPI is composed of both a \textbf{simulation cost} --- which corresponds to the time spent making Monte Carlo Simulation using the current policy --- and a \textbf{learning cost} which corresponds to the time spent learning the classifier that will define the next policy\footnote{In practice, when there are many actions, simulation cost is significantly higher than learning cost, which is thus ignored \cite{Laz10}.}. 
This cost takes the following general form:
\begin{equation}
 Cost = \numstates  \numactions  \numrollouts \times \trajlength \mathcal{I}(\numactions)  + \mathcal{C}(\numstates,\numactions),
\end{equation}
where $\trajlength \mathcal{I}(\numactions)$ is the cost of sampling one trajectory of size $\trajlength$, $\numstates \numactions \numrollouts \times  \trajlength \mathcal{I}(\numactions)$ is the cost of executing the $\numrollouts$ Monte Carlo Simulations over $\numstates$ states testing $\numactions$ possible actions, and $\mathcal{C}(\numstates,\numactions)$ is the cost of learning the corresponding classifier\footnote{We do not consider the computational cost of transitions in the MDP.}. 

The main difference between RCPI and ERCPI comes from the values of $\mathcal{I}(\numactions)$ and $\mathcal{C}(\numstates,\numactions)$. When comparing ERCPI with a RCPI algorithm using a \textit{one-vs-all} (RCPI-OVA) multiclass classifier --- one binary classifier learned for each possible action --- it is easy to see that our method reduces both $\mathcal{I}(\numactions)$ and $\mathcal{C}(\numstates,\numactions)$ by a factor of $\frac{\numactions}{\log \numactions}$ --- cf. Table \ref{table:cost}.

\begin{table} 
\small{
\begin{center}\begin{tabular}{|c||c|c|}\hline
Algorithm & Simulation Cost & Learning Cost \\ \hline \hline
RCPI-OVA & $\numstates  \numactions  \numrollouts ( \trajlength  \numactions)$ & $\numactions. \mathcal{C}(\numstates)$ \\ \hline
ERCPI & $\numstates  \numactions  \numrollouts ( \trajlength \numclassifierslog)  $& $\numclassifierslog. \mathcal{C}(\numstates) $\\ \hline
BRCPI & $\numclassifierslog \left( 2\numstates\numrollouts ( 2\trajlength ) \right) $ & $\numclassifierslog \mathcal{C}(\numstates) $ \\ \hline
\end{tabular}
\end{center}
}
\caption{Cost of one iteration of RCPI OVA, ERCPI, and BRCPI. $\numstates$ is the number of states, $\numactions$ the number of actions, $\numrollouts$ the number of rollouts, $\trajlength$ is trajectory length, $\mathcal{C}(\numstates, \numactions)$ is the cost of learning a classifier for $\numstates$ states, $\numactions$ actions.}
\label{table:cost}
\end{table}

\begin{table}
\small{
\begin{center}
\begin{tabular}{|c||c|c|c|} \hline
Method & RCPI OVA & ERCPI & BRCPI \\ \hline
Complexity & $\mathcal{O}(\numactions^2)$ & $\mathcal{O}(\numactions \log(\numactions))$ & $\mathcal{O}(\log(\numactions))$ \\ \hline
\end{tabular}
\end{center}
}
\caption{Complexity w.r.t. the number of possible actions.}
\label{table:compl}
\end{table}

When considering the BRCPI algorithm, $\mathcal{I}$ and $\mathcal{C}$ are reduced as in ERCPI.  However, the simulation cost is reduced as well, as our method proposes to learn a set of optimal binary policies on $\gamma \log(A)$ binary sub-MDPs. For each of these sub-problems, the simulation cost is $ 2\numstates\numrollouts ( 2\trajlength )$ since the number of possible actions is only 2. The learning cost corresponds to learning only $\gamma \log(A)$ binary classifiers resulting in a very low cost --- cf. Table \ref{table:cost}. The overall resulting complexity w.r.t. to the number of actions is presented in Table \ref{table:compl}, showing that the complexity of BRCPI is only logarithmic. In addition, it is important to note that each of the BRCPI sub-problems is atomic, and are therefore easily parallelized. To illustrate these complexities, computation times are reported in the experimental section.

%% file: sections/experiments.tex
In this paper, our concern is really about being able to deal with a large number of uncorrelated actions in practice. Hence, the best demonstration of this ability is to provide an experimental assessment of ERCPI and BRCPI. In this section, we show that BRCPI exhibits very important speed-ups, turning days of computations into hours or less.

\subsection{Protocol}

We evaluate our approaches on two baseline RL problems: Mountain Car and Maze. 

\begin{figure}[ht]
\begin{center}
\vspace{-1cm}\includegraphics[width=1.0\linewidth]{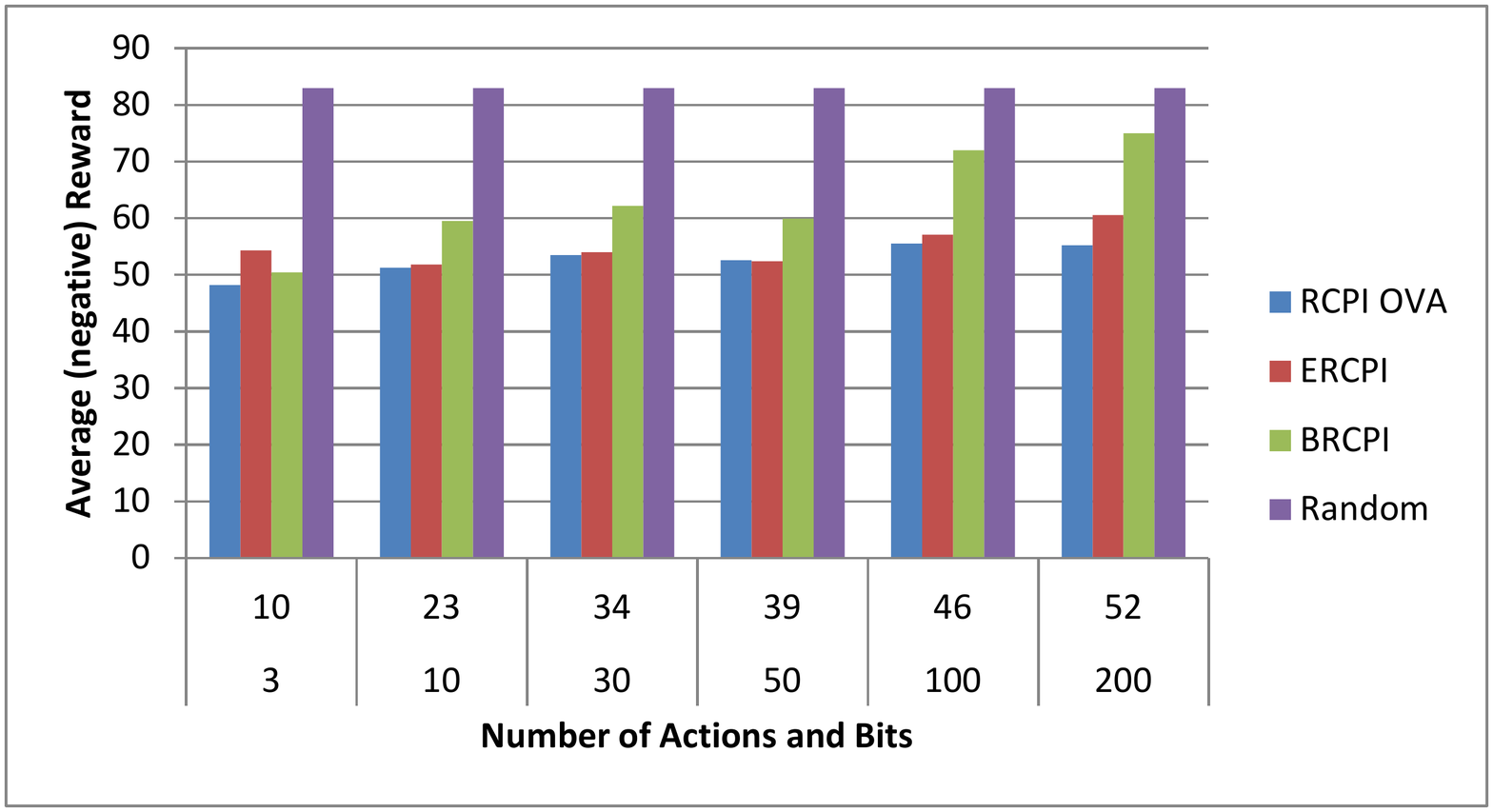}\vspace{-1cm}
\end{center}
\caption{\textbf{Mountain Car}: Average reward (negative value: the smaller, the better) obtained by the different algorithms on 3 runs with different numbers of actions. On the X-axis, the first line corresponds to $\gamma \log(\numactions)$ while the second line is the number of actions $\numactions$.}
\label{fig:mc}
\end{figure}

The first problem, \textbf{Mountain Car}, is well-known in the RL community. Its definition varies, but it is usually based on a discrete and small set of actions (2 or 3). However, the actions may be defined over a continuous domain, which is more ``realistic''. In our experiment, we discretize the range of accelerations to obtain a discrete set of actions. Discretization ranges from coarse to fine in the experiments, thus allowing us to study the effect of the size of the action set on the performance of our algorithms. The continuous \textit{state} space is handled by way of tiling \cite{Sutton1996}. The reward at each step is -1, and each episode has a maximum length of 100 steps.  The overall reward thus measures the ability of the obtained policy to push the car up to the mountain quickly.

The second problem, \textbf{Maze}, is a 50x50 grid-world problem in which the learner has to go from the left side to the right side of a grid. Each cell of the grid corresponds to a particular negative reward, either $-1$, $-10$, or $-100$. For the simplest case, the agent can choose either to move \textit{up}, \textit{down}, or \textit{right}, resulting in a 3-action MDP. We construct more complex action sets by generating all sequences of actions of a defined length i.e. for length 2, the 6 possible actions are \textit{up-up}, \textit{up-right}, \textit{down-up}, etc. Contrary to Mountain Car, there is no notion of similarity between actions in this maze problem w.r.t. their consequences. Each state is represented by a vector of features that contains the information about the different types of cells that are contained in a 5x5 grid around the agent. The overall reward obtained by the agent corresponds to its ability to go from the left to the right, avoiding cells with high negative rewards. 


\begin{figure}[ht]
\begin{center}
\includegraphics[width=1.0\linewidth]{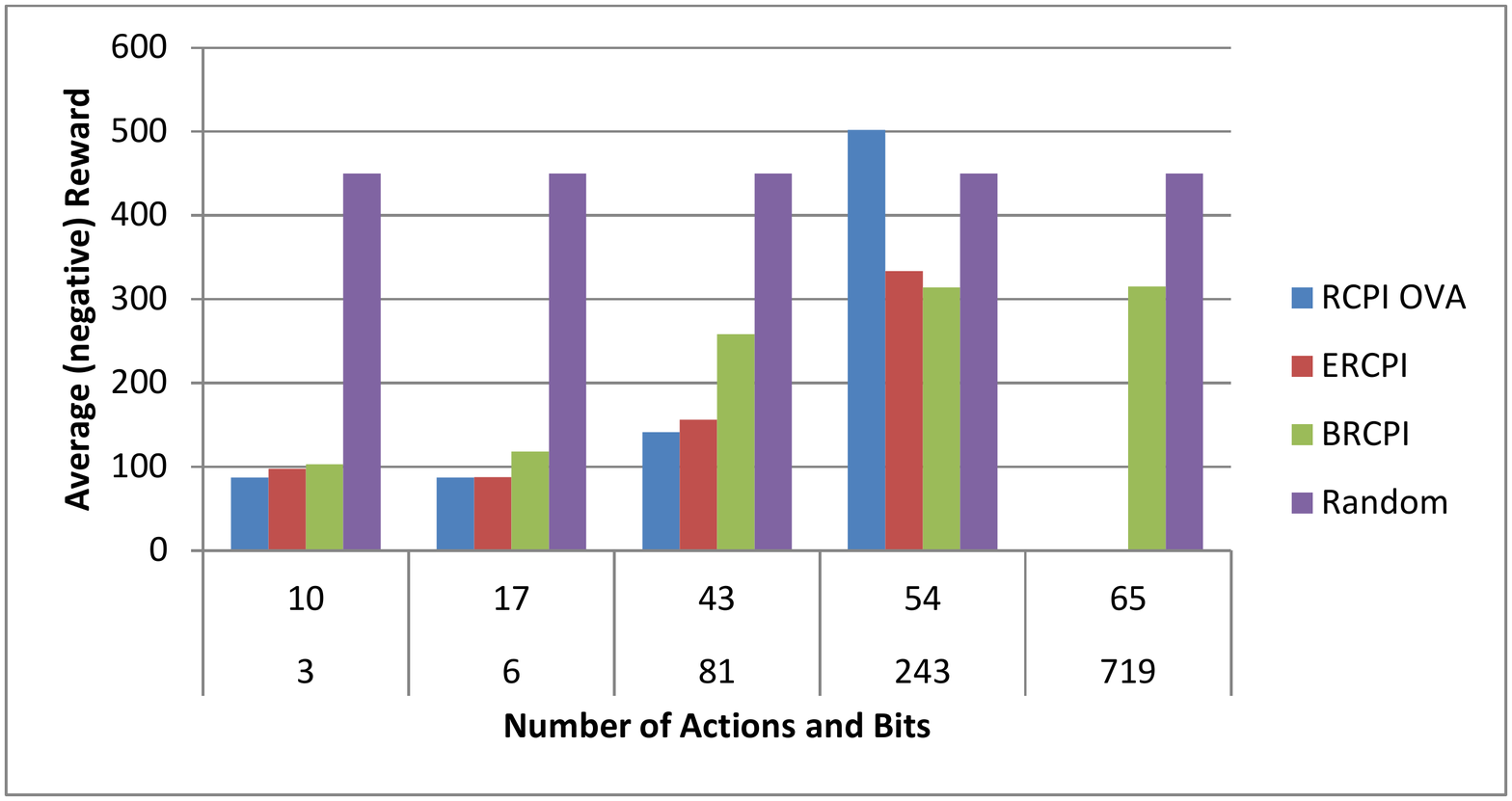}\vspace{-1cm}
\end{center}
\caption{\textbf{Maze}: Average reward (negative value: the smaller, the better) obtained by the different algorithms on 3 different random mazes with different numbers of actions. On the X-axis, the first line corresponds to $\gamma \log(\numactions)$ while the second line is the number of actions $\numactions$.  OVA and ERCPI were intractable for 719 actions.  Note that for 243 actions, RCPI-OVA learns a particularly bad policy.}
\label{fig:maze}
\end{figure}

In both problems, training and testing states are sampled uniformly in the space of the possible states. We have chosen to sample $\numstates=1000$ states for each problem, the number of trajectories made for each state-action pair is $\numrollouts=10$. The binary base learner is a hinge-loss perceptron learned with 1000 iterations by stochastic gradient-descent algorithm. The error correcting codes have been generated using a classical random procedure as in \cite{berger1999error}. The $\alpha$-value of the alpha-mixture policy is $0.5$.

\subsection{Results}
The average rewards obtained after convergence of the three algorithms are presented in Figures \ref{fig:maze} and \ref{fig:mc} with a varying number of actions. The average reward of a random policy is also illustrated. First of all, one can see that RCPI-OVA and ERCPI perform similarly on both problems except for Maze with 243 actions. This can be explained by the fact that OVA strategies are not able to deal with problems with many classes when they involve solving binary classification problems with few positive examples. In this setting, ECOC-classifiers are known to perform better.  BRCPI achieves lower performances than OVA-RCPI and ERCPI. Indeed, BRCPI learns optimal independent binary policies that, when used together, only correspond to a sub-optimal overall policy. Note that even with a large number of actions, BRCPI is able to learn a relevant policy --- in particular, Maze with 719 actions shows BRCPI is clearly better than the random baseline, while the other methods are simply intractable. This is a very interesting result since it implies that BRCPI is able to find non-trivial policies when classical approaches are intractable.

Table \ref{tab:time} provides the computation times for one iteration of the different algorithms for Mountain Car with 100 actions.  ERCPI speeds-up RCPI by a factor 1.4 while BRCPI is 12.5 times faster than RCPI, and 23.5 times faster when considering only the simulation cost. This explains why Figure \ref{fig:maze} does not show performances obtained by RCPI and ERCPI on the maze problem with 719 actions: in that setting, one iteration of these algorithms takes days while only requiring a few hours with BRCPI. Note that these speedup values increase with the number of actions.

\begin{figure}[h]
\includegraphics[width=1.0\linewidth]{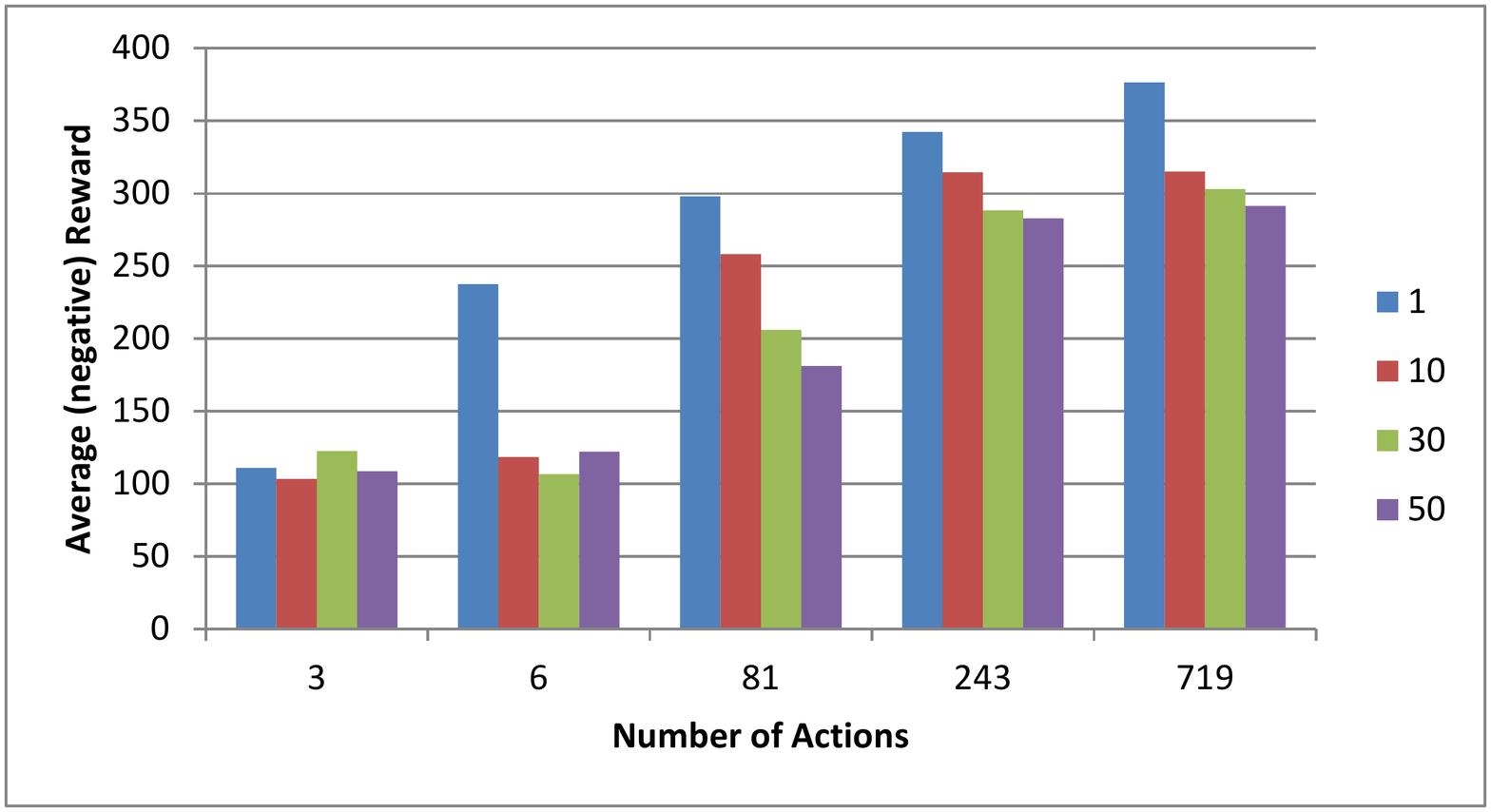}\vspace{-1cm}
\caption{\textbf{Maze Rollouts}: Average reward (negative value: the smaller, the better) obtained by BRCPI for $K = 1,10,30,50$.}
\label{fig:mazemaze}
\end{figure}

At last, Figure \ref{fig:mazemaze} gives the performance of BRCPI depending on the number of rollouts, and shows that a better policy can be found by increasing the value of $K$. Note that, even if we use a large value of $K$, BRCPI's running time remains low w.r.t. to OVA-RCPI and ERCPI.

\begin{table}[ht]
\begin{tabular}{|c|cccc|} \hline
\multicolumn{5}{|c|}{Mountain Car - 100 Actions - 46 bits} \\ \hline \hline
 & Sim. & Learning & Total & Speedup \\ \hline
OVA   & 4,312 & 380 & 4,698 & $\times1.0$   \\ 
ERCPI & 3,188 & 190 & 3,378 & $\times1.4 (\times1.35)$ \\
BRCPI & 184 & 190 & 374 & $\times12.5 (\times23.5)$ \\ \hline
\end{tabular}
\caption{Time (in seconds) spent for one iteration --- during simulation and learning --- of the different variants of the RCPI algorithms using a Xeon-X5690 Processor and a TESLA M2090 GPU for $K=10$ and $S=1000$. The total speedup (and simulation speedup) w.r.t. OVA-RCPI are presented on the last column.}
\label{tab:time}
\end{table}

%% file: sections/related.tex
Rollout Classification Policy Iteration \cite{Lagoudakis2003} provides an algorithm for RL in MDPs that have a very large state space.  RCPI's Monte-Carlo sampling phase can be very costly, and a couple approaches have been provided to better sample the state space \cite{Dimitrakakis2008}, thus leading to speedups when using RCPI. Recently, the effectiveness of RCPI has been theoretically assessed \cite{Laz10}.
The well known efficiency of this method for real-world problems and its inability to deal with many actions have motivated this work. 
\linebreak
Reinforcement Learning has long been able to scale to state-spaces with many (if infinite) states by generalizing the value-function over the state space \cite{tham1994modular, Tesauro92practicalissues}. Tesauro first introduced rollouts \cite{Tesauro1997}, leveraging Monte-Carlo sampling for exploring a large state and action space. Dealing with large action spaces has additionally been considered through sampling or gradient descent on $Q$ \cite{ negoescu2010knowledge, Lazaric2007}, but these approaches assume a well-behaved $Q$-function, which is hardly guaranteed.

There is one vein of work reducing action-space look-ups logarithmically by imposing some form of binary search over the action space \cite{Pazisb, Pazis2011}. These approaches augment the MDP with a structured search over the action space, thus placing the action space's complexity in the state space. Although not inspirational to ERCPI, these approaches are similar in their philosophy. However, neither proposes a solution to speeding up the learning phase as BRCPI does, nor do they eschew value functions by relying solely on classifier-based approaches as ERCPI does.
\linebreak
Error-Correcting Output Codes were first introduced by Dietterich and Bakiri (\citeyear{Dietterich1995}) for use in the case of multi-class classification.  Although not touched upon in this article, coding dictionary construction can be a key element to the ability of the ECOC-based classifier's abilities\cite{Beygelzimera}.  Although in our case we rely on randomly generated codes, codes can be learned from the actual training data \cite{Crammer2002} or from an \textit{a priori} metric upon the classes space or a hierarchy \cite{Cisse2011}.

%% file: sections/conclusion.tex
We have proposed two new algorithms which aim at obtaining a good policy while learning faster than the standard RCPI algorithm.  ERCPI is based on the use of Error Correcting Output Codes with RCPI, while BRCPI consists in decomposing the original MDP in a set of binary-MDPs which can be learned separately at a very low cost. While ERCPI obtains equivalent or better performances than the classical \textit{One Vs. All} RCPI implementations at a lower computation cost, BRCPI allows one to obtain a sub-optimal policy very fast, even if the number of actions is very large. The complexity of the proposed solutions are $\mathcal{O}(\numactions \log(\numactions))$ and $\mathcal{O}( \log(\numactions))$ respectively, in comparison to RCPI's complexity of $\mathcal{O}(\numactions^2)$. Note that one can use BRCPI to discover a good policy, and then ERCPI in order to improve this policy; this practical solution is not studied in this paper. 

This work opens many new research perspectives: first, as the performance of BRCPI directly depends on the quality of the codes generated for learning, it can be very interesting to design automatic methods able to find the well-adapted codes, particularly when one has a metric over the set of possible actions. From a theoretical point of view, we plan to study the relation between the performances of the sub-policies $\pi_i$ in BRCPI and the performance of the final obtained policy $\pi$. At last, the fact that our method allows one to deal with problems with thousands of discrete actions also opens many applied perspectives, and can allow us to find good solutions for problems that have never been studied before because of their complexity.

%% file: main.bbl
\begin{thebibliography}{18}
\providecommand{\natexlab}[1]{#1}
\providecommand{\url}[1]{\texttt{#1}}
\expandafter\ifx\csname urlstyle\endcsname\relax
  \providecommand{\doi}[1]{doi: #1}\else
  \providecommand{\doi}{doi: \begingroup \urlstyle{rm}\Url}\fi

\bibitem[Bentley(1975)]{Bentley1975}
Bentley, Jon~Louis.
\newblock {Multidimensional binary search trees used for associative
  searching}.
\newblock \emph{Communications of the ACM}, 18\penalty0 (9):\penalty0 509--517,
  1975.

\bibitem[Berger(1999)]{berger1999error}
Berger, A.
\newblock Error-correcting output coding for text classification.
\newblock In \emph{Workshop on Machine Learning for Information Filtering,
  IJCAI '99}, 1999.

\bibitem[Beygelzimer et~al.(2008)Beygelzimer, Langford, and
  Zadrozny]{Beygelzimera}
Beygelzimer, A., Langford, J., and Zadrozny, B.
\newblock Machine learning techniques—reductions between prediction quality
  metrics.
\newblock \emph{Performance Modeling and Engineering}, pp.\  3--28, 2008.

\bibitem[Bubeck et~al.(2011)Bubeck, Munos, Stoltz, Szepesv{\'a}ri,
  et~al.]{bubeck2011x}
Bubeck, S., Munos, R., Stoltz, G., Szepesv{\'a}ri, C., et~al.
\newblock X-armed bandits.
\newblock \emph{Journal of Machine Learning Research}, 12:\penalty0 1655--1695,
  2011.

\bibitem[Ciss\'{e} et~al.(2011)Ciss\'{e}, Artieres, and Gallinari]{Cisse2011}
Ciss\'{e}, M., Artieres, T., and Gallinari, Patrick.
\newblock {Learning efficient error correcting output codes for large
  hierarchical multi-class problems}.
\newblock In \emph{Workshop on Large-Scale Hierarchical Classification
  ECML/PKDD '11}, pp.\  37--49, 2011.

\bibitem[Crammer \& Singer(2002)Crammer and Singer]{Crammer2002}
Crammer, Koby and Singer, Yoram.
\newblock {On the Learnability and Design of Output Codes for Multiclass
  Problems}.
\newblock \emph{Machine Learning}, 47\penalty0 (2):\penalty0 201--233, 2002.

\bibitem[Dietterich \& Bakiri(1995)Dietterich and Bakiri]{Dietterich1995}
Dietterich, T.G. and Bakiri, G.
\newblock {Solving multiclass learning problems via error-correcting output
  codes}.
\newblock \emph{Jo. of Art. Int. Research}, 2:\penalty0 263–286, 1995.

\bibitem[Dimitrakakis \& Lagoudakis(2008)Dimitrakakis and
  Lagoudakis]{Dimitrakakis2008}
Dimitrakakis, Christos and Lagoudakis, Michail~G.
\newblock {Rollout sampling approximate policy iteration}.
\newblock \emph{Machine Learning}, 72\penalty0 (3):\penalty0 157--171, July
  2008.

\bibitem[Lagoudakis \& Parr(2003)Lagoudakis and Parr]{Lagoudakis2003}
Lagoudakis, Michail~G. and Parr, Ronald.
\newblock {Reinforcement learning as classification: Leveraging modern
  classifiers}.
\newblock In \emph{Proc. of ICML '03}, 2003.

\bibitem[Lazaric et~al.(2007)Lazaric, Restelli, and Bonarini]{Lazaric2007}
Lazaric, Alessandro, Restelli, Marcello, and Bonarini, Andrea.
\newblock {Reinforcement Learning in Continuous Action Spaces through
  Sequential Monte Carlo Methods}.
\newblock In \emph{Proc. of NIPS '07}, 2007.

\bibitem[Lazaric et~al.(2010)Lazaric, Ghavamzadeh, and Munos]{Laz10}
Lazaric, Alessandro, Ghavamzadeh, Mohammad, and Munos, R{\'e}mi.
\newblock Analysis of a classification-based policy iteration algorithm.
\newblock In \emph{Proc. of ICML '10}, pp.\  607--614, 2010.

\bibitem[Negoescu et~al.(2011)Negoescu, Frazier, and
  Powell]{negoescu2010knowledge}
Negoescu, D.M., Frazier, P.I., and Powell, W.B.
\newblock The knowledge-gradient algorithm for sequencing experiments in drug
  discovery.
\newblock \emph{INFORMS J. on Computing}, 23\penalty0 (3):\penalty0 346--363,
  2011.

\bibitem[Pazis \& Lagoudakis(2011)Pazis and Lagoudakis]{Pazisb}
Pazis, Jason and Lagoudakis, Michail~G.
\newblock {Reinforcement Learning in Multidimensional Continuous Action
  Spaces}.
\newblock In \emph{Proc. of Adaptive Dynamic Programming and Reinf. Learn.},
  pp.\  97--104, 2011.

\bibitem[Pazis \& Parr(2011)Pazis and Parr]{Pazis2011}
Pazis, Jason and Parr, Ronald.
\newblock {Generalized Value Functions for Large Action Sets}.
\newblock In \emph{Proc. of ICML '11}, pp.\  1185--1192, 2011.

\bibitem[Sutton(1996)]{Sutton1996}
Sutton, RS.
\newblock {Generalization in reinforcement learning: Successful examples using
  sparse coarse coding}.
\newblock In \emph{Proc. of NIPS '96}, pp.\  1038--1044, 1996.

\bibitem[Tesauro(1992)]{Tesauro92practicalissues}
Tesauro, Gerald.
\newblock Practical issues in temporal difference learning.
\newblock \emph{Machine Learning}, 8:\penalty0 257--277, 1992.

\bibitem[Tesauro \& Galperin(1997)Tesauro and Galperin]{Tesauro1997}
Tesauro, Gerald and Galperin, Gregory~R.
\newblock {On-Line Policy Improvement Using Monte-Carlo Search}.
\newblock In \emph{Proc. of NIPS '97}, pp.\  1068--1074, 1997.

\bibitem[Tham(1994)]{tham1994modular}
Tham, C.K.
\newblock \emph{Modular on-line function approximation for scaling up
  reinforcement learning}.
\newblock PhD thesis, University of Cambridge, 1994.

\end{thebibliography}
